\documentclass[letterpaper, 10 pt, conference]{ieeeconf}
\IEEEoverridecommandlockouts %
\usepackage[utf8]{inputenc} %
\usepackage[T1]{fontenc}    %
\usepackage[english]{babel} %

\usepackage{overpic}
\usepackage{color}
\usepackage[nice]{nicefrac}
\usepackage{algpseudocode}
\usepackage[linesnumbered,ruled,vlined]{algorithm2e}

\SetCommentSty{mycommfont}
\usepackage{algpseudocode}
\usepackage{multirow}
\usepackage{adjustbox}
\usepackage{booktabs}
\usepackage{amsfonts}
\usepackage{amssymb}
\usepackage[font=small,labelfont=bf]{caption}
\usepackage{subcaption}
\usepackage{amsmath}
\usepackage{cancel}
\usepackage{dsfont}

\usepackage[pagebackref=false,breaklinks=true,colorlinks,urlcolor=blue,citecolor=blue,linkcolor=blue,bookmarks=false]{hyperref}
\usepackage[capitalize, noabbrev]{cleveref}

\makeatletter
\let\NAT@parse\undefined
\makeatother
\usepackage[numbers,sort]{natbib}

\definecolor{turquoise}{cmyk}{0.65,0,0.1,0.3}
\definecolor{purple}{rgb}{0.65,0,0.65}
\definecolor{dark_green}{rgb}{0, 0.7, 0}
\definecolor{orange}{rgb}{0.8, 0.6, 0.2}
\definecolor{red}{rgb}{0.8, 0.2, 0.2}
\definecolor{darkred}{rgb}{0.6, 0.1, 0.05}
\definecolor{blueish}{rgb}{0.0, 0.3, .6}
\definecolor{light_gray}{rgb}{0.7, 0.7, .7}
\definecolor{pink}{rgb}{1, 0, 1}
\definecolor{greyblue}{rgb}{0.25, 0.25, 1}

\let\labelindent\relax
\usepackage{enumitem}
\setlist[itemize]{noitemsep,leftmargin=*,topsep=0in}
\setlist[enumerate]{noitemsep,leftmargin=*,topsep=0in}

\newcommand{\at}[1]{{\color{blueish}#1}} %

\newcommand{\llg}[1]{{\color{dark_green}#1}}

\renewcommand{\at}[1]{{\color{black}#1}}
\renewcommand{\llg}[1]{{\color{black}#1}}

\newcommand{\CIRCLE}[1]{\raisebox{.5pt}{\footnotesize \textcircled{\raisebox{-.6pt}{#1}}}}

\newcommand{\Figure}[1]{Figure~\ref{fig:#1}}

\newcommand{\eq}[1]{(\ref{eq:#1})}

\newcommand{\Section}[1]{Section~\ref{sec:#1}}

\usepackage{blindtext}
\usepackage{bbm}

\usepackage{lipsum}
\usepackage[]{microtype}

\renewcommand{\paragraph}[1]{\vspace{.5em}\noindent\textit{#1} --}

\newcommand{\algoName}{nerf2nerf\xspace}

\DeclareMathOperator*{\argmin}{arg\,min}

\newcommand{\loss}[1]{\mathcal{L}_\text{#1}}
\newcommand{\expect}{\mathbb{E}}
\newcommand{\real}{\mathbb{R}}

\newcommand{\rfield}{\mathcal{R}}

\renewcommand{\S}{\mathcal{S}}
\newcommand{\x}{\mathbf{x}}
\newcommand{\viewdir}{\mathbf{d}}
\newcommand{\ball}{\mathcal{B}}
\newcommand{\object}{\mathcal{X}}
\newcommand{\pose}{\mathbf{T}}

\newcommand{\transmittance}{\mathcal{T}}
\newcommand{\zero}{\mathbf{0}}
\newcommand{\SE}{\text{SE}}
\renewcommand{\time}{{(t)}}
\newcommand{\keypoints}{\mathcal{Q}}
\newcommand{\keypoint}{\mathbf{q}}
\newcommand{\Dirs}{\mathcal{D}}
\newcommand{\kernelCutoff}{c}
\newcommand{\kernelSlope}{\alpha}
\newcommand{\activeset}{\mathcal{A}}
\newcommand{\R}{\mathbf{R}}
\renewcommand{\t}{\mathbf{t}}

\newcommand{\var}{\sigma}

\newcommand{\normal}{\mathcal{N}}
\newcommand{\posenc}{\gamma}
\newcommand{\cov}{\Sigma}

\newcommand{\resolution}{\rho}

\newcommand{\image}{\mathbf{I}}
\newcommand{\C}{\boldsymbol{C}} %
\newcommand{\gt}{\text{gt}} %

\newcommand{\params}{\boldsymbol{\theta}}

\newcommand{\radiance}{\mathbf{c}}

\newcommand{\ray}{\mathbf{r}}
\newcommand{\origin}{\mathbf{o}}
\newcommand{\dir}{\mathbf{d}}

\newcommand{\density}{\tau}

\newcommand{\given}{|}

\newcommand{\kernel}{\kappa}

\newcommand{\diff}{dt}
\newcommand{\threshold}{\epsilon}
\newcommand{\surface}{\mathcal{S}}
\newcommand{\candidates}{\mathcal{C}}

\title{\LARGE \bf
\algoName: Pairwise Registration of Neural Radiance Fields
}

\author{
Lily Goli$^{1, 2}$, \, %
Daniel Rebain$^{4}$, \, %
Sara Sabour$^{1,2,6}$, \, %
Animesh Garg$^{1,2,5}$, \, %
Andrea Tagliasacchi$^{1,3,6}$
\\[-.5em]
\\[.1em]
$^{1}$University of Toronto,
$^{2}$Vector Institute,
$^{3}$SFU,
$^{4}$UBC,
$^{5}$NVIDIA,
$^{6}$Google Research
}

\begin{document}

\thispagestyle{empty}
\pagestyle{empty}

\twocolumn[{%
\renewcommand\twocolumn[1][]{#1}%
\maketitle
\begin{center}
\centering
\captionsetup{type=figure}
\begin{overpic} 
    [width=.99\linewidth]
    {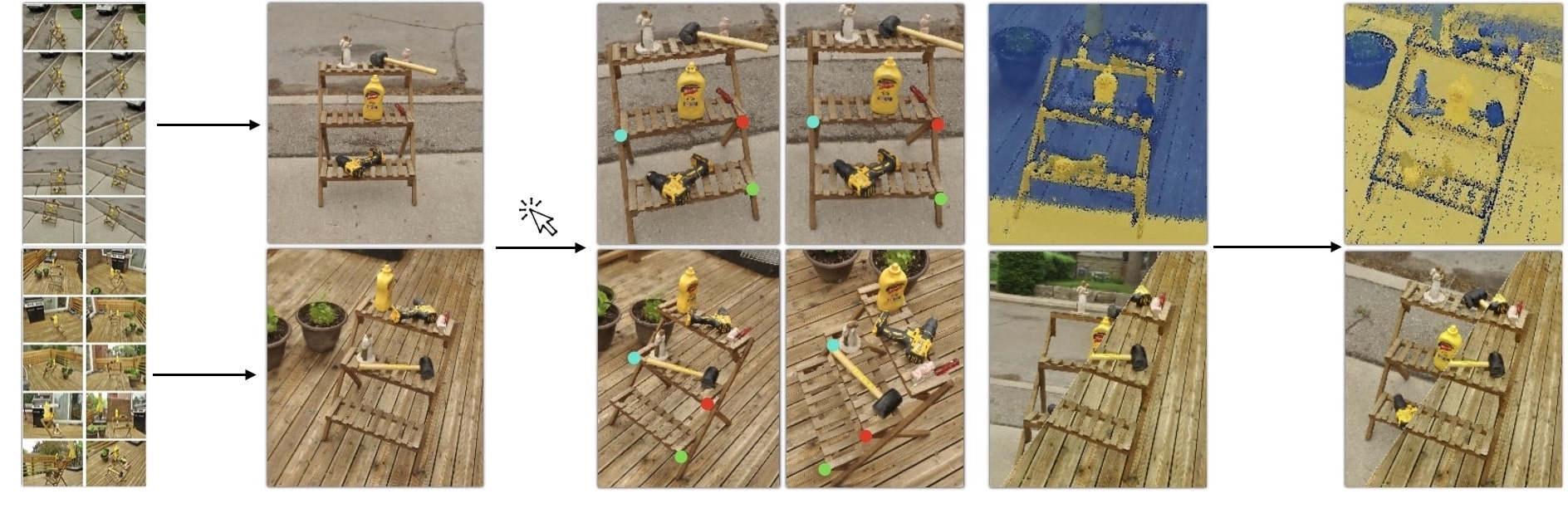}
    \put(-0.7,22){\rotatebox{90}{Scene A}}
    \put(-0.7,7){\rotatebox{90}{Scene B}}
    
    \put(10.,11.5){\parbox{3em}{\centering \small Train\\ NeRF}}
    \put(10.,27.5){\parbox{3em}{\centering \small Train\\ NeRF}}
    
    \put(1.8,-0.2){\parbox{4em}{\centering \small Input\\ Images}}
    \put(17,-0.2){\parbox{7em}{\centering \small Neural\\ Radiance Fields}}
    \put(44.7,-0.2){\parbox{5em}{\centering \small Keypoint\\ Annotations}}
    \put(65,-0.2){\parbox{5em}{\centering \small Keypoint\\ Registration}}
    \put(88,-0.2){\parbox{5em}{\centering \small Optimized\\ Registration}}
    
    \put(76.5,19){\parbox{5em}{\centering \tiny $r(\S_a,\pose(\S_b))$}}
    \put(76.7,16){\parbox{5em}{\centering \small $\argmin$}} %
    \put(76.7,14){\parbox{5em}{\centering \small $\pose$}}
\end{overpic}
\captionof{figure}{\textbf{Teaser} -- 
\at{We propose a novel method for 3D registration that operates directly on Neural Radiance Fields (NeRFs) that have been pre-trained from image collections.
Specifically, we seek to solve the task of aligning partially overlapping geometry from two~(different) scenes, given only a few human annotations to constrain the solution to a single valid registration (out of multiple possible alignments).
Registering sparse keypoints leads to only an \textit{approximate} registration, hence we define an optimization to precisely align the fields to each other.
To achieve this, we introduce the \textit{surface field} $\surface$ as a geometric representation that can be extracted from a NeRF, in lieu of converting NeRF to classical geometric representations like point clouds or polygonal meshes.}
}
\label{fig:teaser}
\vspace{-5pt}
\end{center}
}]

\begin{abstract}
We introduce a technique for pairwise registration of neural fields that extends classical optimization-based local registration (i.e. ICP) to operate on Neural Radiance Fields~(NeRF)~--~neural 3D scene representations trained from collections of calibrated images.
NeRF does not decompose illumination and color, so to make registration invariant to illumination, we introduce the concept of a ``surface field''~--~a field distilled from a pre-trained NeRF model that measures the likelihood of a point being on the surface of an object.
We then cast nerf2nerf registration as a robust optimization that iteratively seeks a rigid transformation that aligns the surface fields of the two scenes.
We evaluate the effectiveness of our technique by introducing a dataset of pre-trained NeRF scenes~--~our synthetic scenes enable quantitative evaluations and comparisons to classical registration techniques, while our real scenes demonstrate the validity of our technique in real-world scenarios. Additional results available at: \url{https://nerf2nerf.github.io}
\end{abstract}

\section{Introduction}
\label{sec:intro}

Pairwise registration of 3D scenes that have been acquired in-the-wild is an essential first step in many computer vision and robotic pipelines such as localization~\cite{localization1, localization2}, pose estimation~\cite{posest1, poset2} and large scene reconstruction~\cite{recons1, recons2}.
Registering 3D scenes that have been acquired in-the-wild presents a number of technical challenges, including noise, outliers, and, most importantly, \textit{partial} overlap~\cite{sparseicp,predator}.

Acquisition of a 3D scene has classically been achieved via photogrammetry~\cite{colmap}, SLAM~\cite{teed2021droid}, or depth-fusion~\cite{kinfu}, where the acquired scene is then represented either in \textit{explicit}~(i.e. a point cloud~\cite{colmap} its reconstructed mesh~\cite{kazhdan2006poisson}) or \textit{implicit} form~(i.e. a truncated signed distance function~\cite{curless1996volumetric}).
Recent work has highlighted how these classical 3D reconstruction pipelines struggle in accurately representing many 3D objects~\cite[Fig.~2]{nerf-supervision}, which is problematic if we rely on them as a vehicle for 3D model-based perception in performing robotics tasks.

However, the recent proliferation of research in \textit{neural fields}~\cite{neural-fields} coupled with the rapid progression of \textit{differentiable rendering}~\cite{neural-rendering}, has created a new class of 3D representations that leverage \textit{volume rendering}~\cite{volren-digest} to relate the underlying 3D substrate to 2D image observations in a differentiable fashion~\cite{neural-surface, mildenhall2020nerf, Niemeyer2020DifferentiableVR}.
Among these, Neural Radiance Fields~(NeRF)~\cite{mildenhall2020nerf} has very rapidly found application in a variety of computer vision applications, and its adoption in robotics has also commenced~\cite{rl-nerf, inerf, nerf-supervision}.

Within less than two years, training time for NeRF models was reduced from \textit{days}~\cite{mildenhall2020nerf} to \textit{seconds}~\cite{instantngp}, and it is not unreasonable to expect that real-time training from streaming video is within reach.
Given these capabilities, one is left to wonder how some of the fundamental tools from 3D geometry processing could be adapted \textit{without} relying on conversion (e.g. NeRF to polygonal mesh) and classical tools (e.g. polygonal mesh processing).
\at{Within this context,} in this paper we consider these fundamental questions:
\begin{enumerate}
\item \textit{``How can a geometric 3D representation be extracted from a pre-trained NeRF?''} \quad such a representation should measure the likelihood of a point in space to lie on the surface of an object -- we introduce the concept of ``surface fields'' as a drop-in replacement for point clouds and polygonal meshes.
\item \textit{``How can two scenes acquired under different illumination be registered?''} \quad
one cannot rely on radiance comparisons to register two scenes, as view-dependent radiance is the compound effect of material~(SVBRDF) and environment lighting -- we exploit surface fields and their invariance to illumination. 
\item \textit{``How can we register two neural fields?''} \quad
while fields as Eulerian in nature, most robust registration algorithms assume a Lagrangian representation -- we introduce, to the best of our knowledge, the first registration technique that achieves this without relying on implicit-to-explicit conversions.
\end{enumerate}

To validate our findings, we introduce a dataset consisting of scenes with partial overlap -- of objects in common between different scenes, and multiple partial observations of an object.
The dataset includes \textit{synthetic} scenes (for quantitative evaluation) as well \textit{real-world} scenes (for qualitative evaluation and to verify applicability to in-the-wild settings).
We evaluate our \textit{\algoName} algorithm on this dataset, and compare its performance to pipelines relying on conversion to classical representations.

\section{Related work}
\label{sec:related}
Classical registration methods between two 3D point clouds can be split into \textit{local} and \textit{global} approaches. 
\quad
\textit{Local} methods are typically a variant of the popular Iterative Closest Points (ICP)~\cite{icp} algorithm -- a method that performs registration by optimizing for zero-length closest-point correspondences.
However, if it is not carefully initialized, ICP registration is known to converge to local minima~\cite{icpv1, icpv2, icpv3, icpv4}.
This makes registration of \textit{partially} overlapping point clouds particularly challenging, as closest-point correspondences can be incorrect even when two partially overlapping scenes are in perfect alignment; this is typically resolved by replacing the least-squares norm with a robust kernel, leading to robust ICP variants~\cite{sparseicp, teaser, Gore, optimalcorr}.
Our solution implements these ideas, but operates on neural fields rather than point clouds, and is capable of registering two pre-trained NeRF models even when they only exhibit partial overlap.
\quad
\textit{Global} methods do not rely on a properly initialized local optimization.
For example, fast global registration (FGR)~\cite{fgr} combines the aforementioned robust optimization with feature matching to register point clouds in partial overlap.
Other global methods such as \cite{super4pcs,BnB, RANSAC, DIEZ-2012}, compare tuples of points between the two point clouds and optimize for transformations via stochastic search.
Recently, neural networks that operate on point clouds~\cite{pointnet,dgcnn} have led to new registration techniques that use deep features for correspondence matching~\cite{pointnetlk, dcp, PRNet,perfMatch}. 

\paragraph{Registration in NeRF}
Registration for NeRFs has mostly been explored from the point of view of \textit{camera pose estimation}.
Such works propose to train a NeRF from a collection of images with noisy/unknown pose and jointly optimize for registration and reconstruction \cite{barf,nerf--, selfcalibrating}. 
The pioneering work iNeRF~\cite{inerf} predicts, via optimization, the camera pose corresponding to an image of the object for which a pre-trained NeRF model is available.
In other words, iNeRF performs \textit{image2nerf} registration, while we perform \textit{nerf2nerf} registration.
Further, while previous methods perform registration by optimizing with losses based on radiance, we solely leverage radiance to extract a geometric representation from each scene.
Consequently, as nerf2nerf relies on the geometry information within the scene, it can register scenes even when there is a mismatch in light configuration.

\subsection{Neural Radiance Fields}
\label{sec:related_nerf}
Neural Radiance Fields \cite{NeRF} (NeRFs) are implicit volumetric representations which encode the appearance and geometry of 3D scenes.
A NeRF model $\rfield$ stores a continuous representation of a 3D scene within the parameters $\params$.
It can be seen as a function that maps a position $\x$ in the scene and the viewing direction $\viewdir$ to a view-dependent radiance $\radiance$ and view-independent density $\density$:
\begin{equation}
\radiance(\x, \dir) ,\:
\density(\x)
= \rfield(\x, \dir ; \params)
\end{equation}
Given a ray $\ray{=}(\origin,\viewdir)$ with origin $\origin$ oriented as $\viewdir$, from which a point can be taken at depth $t$ as $\ray(t){=}\origin {+} t\viewdir$, volume rendering of the radiance field $\radiance(\x, \viewdir)$ is computed as:
\begin{align}
\C(\ray) &= \int_0^t
\transmittance(t; \ray)
\cdot
\density(\ray(t))
\cdot
\radiance(\ray(t), \viewdir) \, dt
\\
\transmittance(t; \ray) &= \exp \left( -\int_0^{t} \density(\ray(s)) \, ds \right)
\end{align}
where $\density(\x)$ is the \textit{volumetric density} (or extinction coefficient, the differential probability of a viewing ray hitting a particle), and $\transmittance(t; \ray)$ is the \textit{transmittance} (or transparency, the probability that the viewing ray travels a distance $t$ along $\ray$ without hitting any particle); see~\cite{volren-digest}.
The network parameters $\theta$ of $\radiance$ and $\density$ are optimized to minimize the squared distance between the predicted color $\C(\ray)$ and ground truth $\C^\gt$ for each ray $\ray$ sampled from image $\image$:
\begin{equation}
\loss{rgb}(\params) = \sum_{i} \expect_{\ray \sim \image_i}
\left[
\| \C(\ray) - \C^\gt_i(\ray) \|_2^2 
\right]
\label{eq:nerfrgb}
\end{equation}
\section{Method}
We are given two 3D scenes represented as (neural) radiance fields, containing a shared substructure $\object$ (e.g. a common object); see~\Figure{teaser}.
We seek a transformation~$\pose{\in}\SE(3)$ that aligns some shared substructure of the two scenes.
Using the notation from~\Section{related_nerf}, these scenes can be represented as radiance fields~$\rfield$:
\begin{align}
\begin{split}
\density_{a}(\x), \radiance_{a}(\x, \viewdir) &= \rfield_{a}(\x, \viewdir), \quad
\x {\in} \ball(\zero, r_{a}), \, \viewdir {\in} \Dirs  \\
\density_{b}(\x), \radiance_{b}(\x, \viewdir) &= \rfield_{b}(\x, \viewdir), \quad
\x {\in} \ball(\zero, r_{b}), \, \viewdir {\in} \Dirs
\end{split}
\label{eq:nerfs}
\end{align}
where $\ball(\zero, r)$ is a Euclidean ball of size $r$ centered at the origin, and $\Dirs$ is the subset of unit vectors in $\real^3$.
Note that, without loss of generality, we assume the fields to be of the same scale and have a bounded domain of radius $r$, centered at the origin.
We make \textit{no further assumptions} about two scenes, such as knowing the extent of the shared substructure, or identical illumination configuration.
Our technique \textit{only assumes} we have access to~$\rfield_{a}$ and $\rfield_{b}$ during optimization~(i.e. the images used to train the NeRF are \textit{not} available).
We formulate the Neural Field registration problem as \textit{iterative} optimization of a two-term energy:
\begin{align}
\argmin_{\pose} \:\:
(1-\lambda^\time) \cdot &\loss{match}(\S_a, \S_b; \pose) &\text{(\Section{matching})}
\label{eq:main}
\\[-.5em]  
+~ \lambda^\time \cdot &\loss{key}(\keypoints_a, \keypoints_b; \pose) 
&\text{(\Section{keypoint})}
\nonumber
\end{align}
where the term $\loss{match}$ (robustly) measures the distance between two fields given the candidate transformation~$\pose$, and $\loss{key}$ measures the distance between two sets of $K$ user-annotated keypoints $\keypoints {\in} \real^{K \times 3}$.
As the registration computed from keypoints alone is typically noisy (see~\Figure{teaser}), the term $\lambda$ anneals the optimization over the  $T{=}10000$ iterations with an \textit{additive trigonometric scheduler}, hence gradually shifting the optimization from registering keypoints with $\loss{key}$ to registering the fields via~$\loss{match}$:
\begin{align}
\lambda^{(t)} = \tfrac{1}{2}\left(1 + \cos{\left(\tfrac{t\pi}{T}\right)}\right).
\label{eq:scheduler}
\end{align}

\paragraph{Light-invariant registration}
As we do not assume the light configuration to be identical across the scenes, we cannot compare the radiance terms in~\eq{nerfs} to determine whether two scenes are well aligned.
Furthermore, working with radiance would require comparisons of \textit{distributions} over~$\Dirs$, as radiance is a view-dependent field.
In other words, we employ NeRF and its radiance formulation to generate a sufficiently accurate 3D representation of the \textit{geometry} within the scene.
Geometry, which in a NeRF model is related to the density field $\density$, is, as desired, lighting invariant.
However, as the values of $\density$ in unobserved areas are not meaningful, in \Section{surfacefield} we introduce the notion of \textit{surface field}, the likelihood of a position in space to lie on the surface of a solid object, and describe how they can be distilled from the density field of a NeRF model.

\paragraph{Sample-efficient optimization}
While in classical registration pipelines~\cite{regcourse} \textit{putative correspondences} are computed via closest-point~\cite{sparseicp} or projective~\cite{kinfu} correspondences, neural fields do not support this mode of operation (i.e. they are not Lagrangian representations, but Eulerian).
One could densely sample $\x \in \ball(\zero, r)$, but this amounts to evaluating the neural model at a very large set of spatial locations.
This is not only computationally expensive, but also pointless, as most of these evaluations will be discarded by the robust kernels implementing $\loss{match}$.
To overcome this problem, we introduce a sampling mechanism based on Metropolis-Hastings sampling; see~\Section{sampling}.

\begin{figure}
\includegraphics[width=.99\columnwidth]{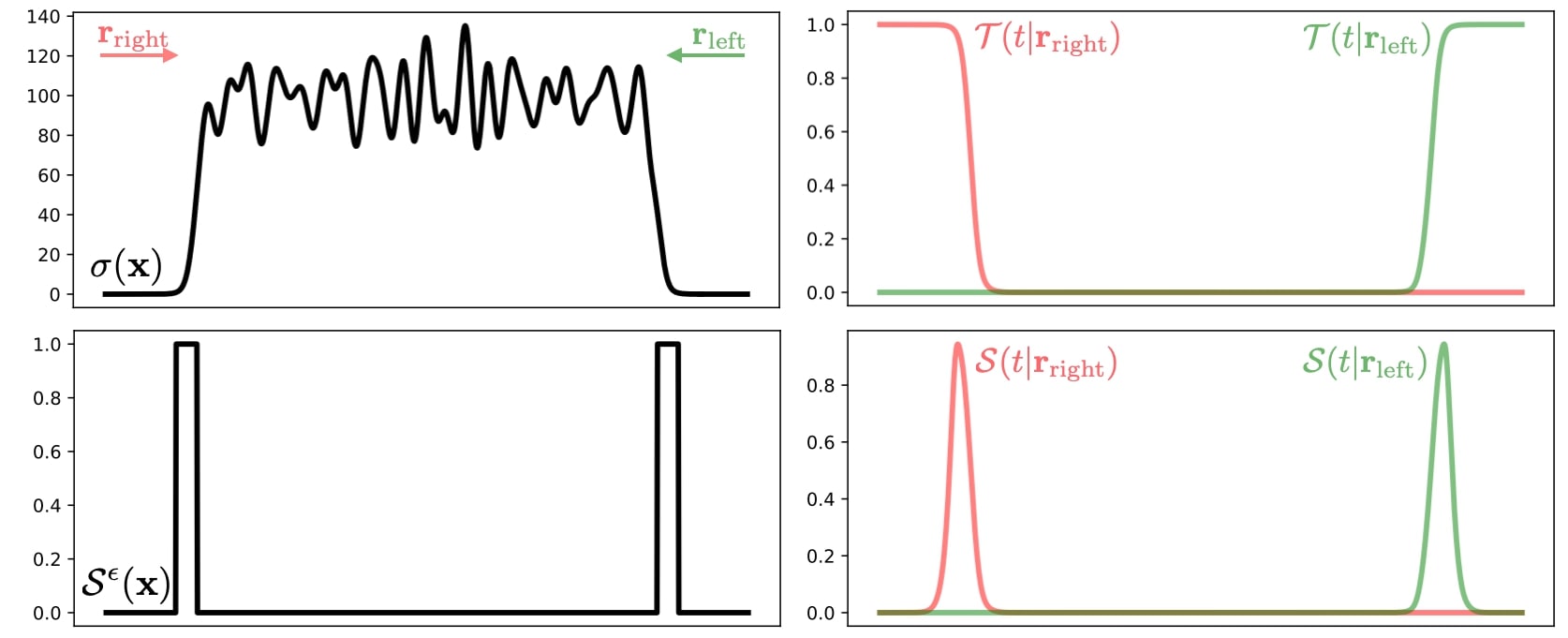}
\begin{overpic} 
[width=.99\columnwidth]{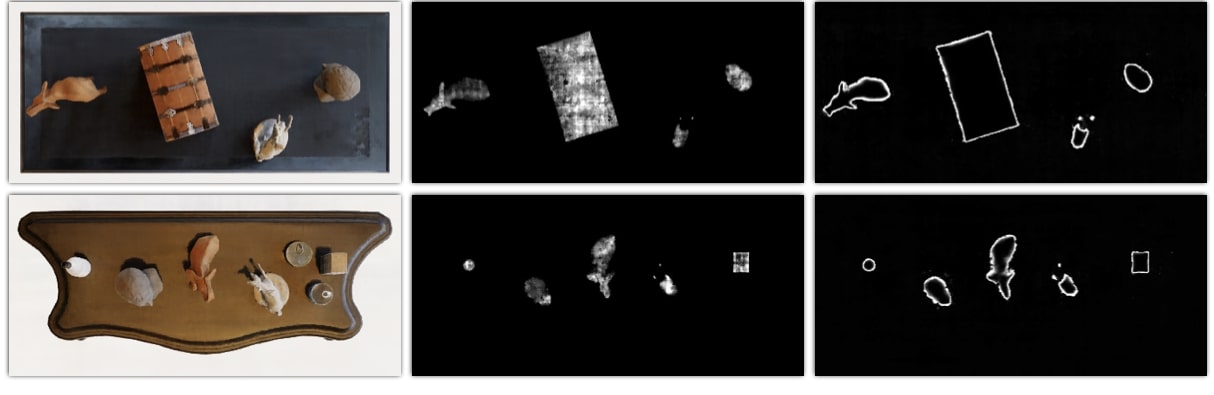}
\put(6,-2.5){Color Image}
\put(39,-2.5){Density Field}
\put(72,-2.5){Surface Field}
\end{overpic}
\caption{
\textbf{Surface field} --
(top) We illustrate the NeRF-like density model for a 2D example of a ``rect`` function observed by two rays: one observing the scene from the left, the other from the right.
NeRF does not supervise density in areas of occlusion, leading to a null-space that often results in a noisy density function.
Transmittance integrates this noise out by modeling occlusion, and our definition of surface field processes the transmittance to reveal the location of the surface.
(bottom) The surface and density fields of two scenes slices at roughly the same height.
While the density field is \textit{not} consistent across scenes, the surface field is.
}
\label{fig:surfacefield}
\end{figure}
\begin{figure*}
\begin{overpic} 
[width=.99\linewidth]
{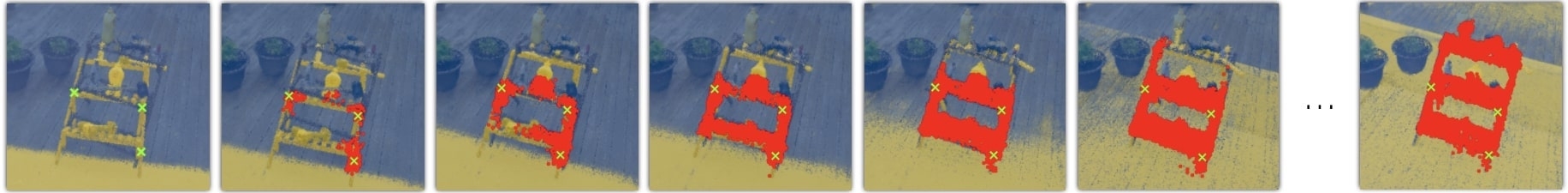}
\put(4.9,-1.5){\footnotesize{$T{=}0$}}
\put(17.7,-1.5){\footnotesize{$T{=}400$}}
\put(31.7,-1.5){\footnotesize{$T{=}800$}}
\put(45,-1.5){\footnotesize{$T{=}1200$}}
\put(58.5,-1.5){\footnotesize{$T{=}1600$}}
\put(72.4,-1.5){\footnotesize{$T{=}2000$}}
\put(90,-1.5){\footnotesize{$T{=}10000$}}

\end{overpic}
\caption{
\textbf{Iterations} --
of registration and visualization of how the sampling set $\activeset^{(t)}$ (depicted by small red dots) changes during optimization; the samples start in the neighborhood of the annotated keypoints (green crosses), and converge to sample the shared portion of space.
}
\label{fig:iterations}
\end{figure*}
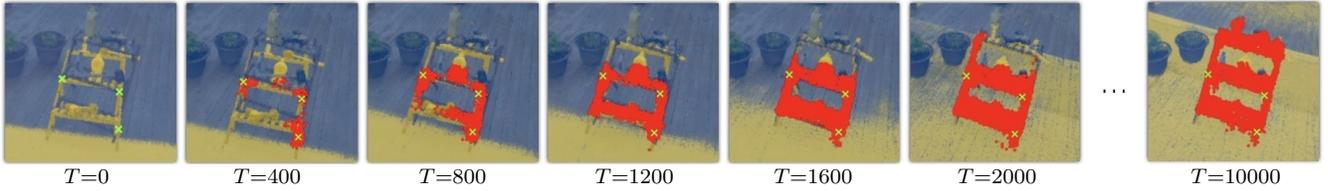
\subsection{Surface Field}
\label{sec:surfacefield}
We seek to define a field that expresses the likelihood of a point being on the surface of an object; see~\Figure{surfacefield}~(top).
Let~$\density(t \given \ray)$ be the \textit{differential probability} of a ray~$\ray{=}(\origin, \dir)$, cast from $\origin$ towards $\dir$, hitting a particle at point~$\ray(t) {=} \origin + t\cdot\dir$; hence, the probability of hitting if we were to move along such ray by an infinitesimal length $\diff$ will be $\density(t \given \ray) \cdot \diff$.
Let the transmittance~$\transmittance(0 {\rightarrow} t \given \ray)$ be the \textit{probability} that a ray $\ray$ hits no solid particle on its way to the point $\ray(t)$.
We can then define the \textit{differential probability} of hitting a surface at point~$\ray(t)$ as the product of the probability of travelling with no obstructions along $\ray$ until position $\ray(t)$ times the differential likelihood of hitting a particle at $\ray(t)$.
We can then locally integrate this differential likelihood in a neighborhood of size $\delta$ around $\ray(t)$ to obtain the likelihood $\surface(t \given \ray)$ of hitting the surface: 
\begin{align}
\surface(t \given \ray) &=
\int_{t-\delta}^{t+\delta} 
\transmittance(0 \rightarrow s \given \ray)
\cdot
\density(s \given \ray)
\; 
ds
\end{align}
where note that density is not a view-dependent quantity like transmittance $\transmittance$, but rather $\density(s \given \ray)\equiv \density(\ray(s))$.
Under the assumption that $\density(s \given \ray)$ has a constant value $\density_t$ in the range $[t-\delta, t+\delta]$, we can prove that (see \cite{volren-digest}):
\begin{align}
\surface(t \given \ray) = \transmittance(0 \rightarrow t-\delta \given \ray) \cdot [ 1 - \exp(-2\density_t\delta)] \in [0,1]
\label{eq:surfacefielddiscrete}
\end{align}
Then, given a point in space~$\x$, and assuming constant density in an Euclidean ball of radius~$\delta$, we can define the surface field as the \textit{maximum} of the likelihoods of hitting a surface given ray travelling from \textit{any} camera through $\x$:
\newcommand{\origins}{\mathcal{O}}
\begin{align}
\surface(\x) &= \max_{\origin{\in}\origins} \:\: \surface \left( \|\origin-\x\| ~\given \left(\origin, \tfrac{\origin-\x}{\|\origin-\x\|}\right) \right) \in [0,1]
\label{eq:surfacefield}
\end{align}
where $\origins$ is the set of camera/ray origins, with one entry from each calibrated image~$\{\image_i\}$.
The density functions between two independently captured scenes might not be perfectly identical, as they are affected by the distribution of cameras, and the lighting configuration.
To robustify the process, we can threshold the field at $\threshold$~(set to $\threshold{=}0.5$ unless otherwise noted) to obtain a \textit{conservative} estimate of the object's surface~(i.e.~a~dilation):
\begin{align}
\surface^{\threshold}(\x) =  \mathds{1}(\surface(\x) > \threshold) \in \{0, 1\}
\label{eq:thresholded}
\end{align}
An example of the surface field for a slice of a 3D scene is visualized in~\Figure{surfacefield}~(bottom).

\subsection{Matching energy}
\label{sec:matching}
The \textit{robust} matching energy compares the two scenes given the \textit{current} estimate of the relative pose $\pose(\x) {=} \R \x {+} \t$:
\begin{align}
\loss{match}(\S_a, \S_b; \pose) &=
\expect_{\x \in \activeset}
\:
\kernel(r(\x; \S_a, \S_b, \pose); \kernelCutoff, \kernelSlope) 
\label{eq:robustmatching}
\end{align}
where  $\activeset$ is an ``active'' set of samples from a sub-portion of~$\ball(0, r_a)$ on which the fields are compared; see~\Section{sampling} for additional details.
\quad
The robust kernel~$\kernel{:}~\real^+ {\rightarrow} \real^+$ makes the registration optimization robust to outliers, and its adaptive hyper-parameters respectively control the decision boundary between inliers/outliers~($\kernelCutoff$) and the impact of outliers in
optimization~($\kernelSlope$); see~\cite{robustkernel}.
\quad
The registration residual~$r(\x; \S_a, \S_b, \pose)$ in~\eq{robustmatching} compares the similarity in the surface field between the two scenes:
\begin{align}
r(\x; \S_a, \S_b, \pose) = \|\S_a^\threshold(\x) -
\S_b^\threshold(\R\x+\t)\|_2
\label{eq:residuals_nondiff}
\end{align}
Differentiating \eq{residuals_nondiff} w.r.t. $\pose$ is challenging due to the fact that~$\S^\threshold(\x)$ is a field whose co-domain is a categorical~$\{0, 1\}$, hence resulting in gradients that are either zero or infinity \llg{(Figure~\ref{fig:surfacefield})}.
We can resolve this issue by convolving the categorical field with a zero-mean Gaussian \llg{of isotropic covariance matrix $\cov {=} \var^2 I$ (unvaried smoothing in all directions)}, as in~\cite{kazhdan2006poisson}:
\begin{equation}
\S^\var(\x) = \S^\threshold(\x) \circledast \normal(0, {\var^2}) 
\approx \expect_{\mathbf{z} \sim \normal(\x,\var^2)} \: \left[ \S^\threshold(\mathbf{z}) \right]
\label{eq:convolution}
\end{equation}
leading to our \textit{differentiable} registration residuals:
\begin{equation}
r(\x; \S_a, \S_b, \pose) = \|\S_a^\sigma(\x) -
\S_b^\sigma(\R\x+\t)\|_2
\end{equation}
where the standard deviation~$\var$, as discussed in \cite[Sec.5.1]{nasa}, controls the \textit{receptive field} of registration: larger values of $\var$ reduce the sensitivity of registration to local minima, while smaller values of $\var$ are critical to achieve high-precision registration.

\paragraph{Distillation}
In contrast to \cite{nasa}, which differentiates through the expectation operator in~\eq{convolution}, we opt for the simpler solution of ``distilling'' the field $\S^\var(\x)$ into a neural network~$\S^\var(\x;\theta)$ with parameters $\theta$.
This is possible because our scenes are \textit{rigid}, and it is beneficial as it results in faster optimization, as fewer neural field executions become necessary; e.g. when querying the same point twice, the evaluation of the expectation in~\eq{convolution} is amortized.\footnote{\at{The performance of this operation could be further accelerated by distilling $\S^\var(\x)$ into a feature grid rather than an MLP, as demonstrated in very recent research~\cite{instantngp,relufields}.}}
We perform this distillation via a conditional neural field implemented through \textit{integrated positional encoding}~\cite{mipnerf}.
Following the derivations from~\cite{mipnerf}, under the assumption of $\cov {=} \var^2 I$, integrating positional encoding represents the point~$\x$ by the (sorted) set:
\begin{align}
\posenc_{\var}(\x) \!=\!
\expect_{\mathbf{z} \in \normal(\x,\var^2)}[\posenc(\mathbf{z})] \!=\!
\left\{
\!\!
\left(
\tfrac{sin(2^l\mathbf{z})\cdot}{\sqrt{exp(4^l\var^2)}}, 
\tfrac{cos(2^l\mathbf{z})\cdot}{\sqrt{exp(4^l\var^2)}}
\right)
\!\!
\right\}_{l=0}^L
\nonumber
\end{align}
where $\posenc(\cdot)$ is positional encoding~\cite{mildenhall2020nerf}, and we employ $L$ frequency bands.
We can then distill the neural field as:
\begin{align}
\argmin_\theta \:
\expect_{\x \in \ball(\zero, r)}
&\left[
\mathcal{L}_\text{poisson}(\S^\var(\posenc_{\var}(\x); \theta)), \: \S^\var(\x)) 
\right]
\label{eq:distill}
\\
&\:\:\mathcal{L}_\text{poisson}(x,y) = x - y \log x 
\end{align}
where the poisson loss function $\mathcal{L}_\text{poisson}$ is used to combat the heavily class-imbalanced distribution of surface fields.

\paragraph{Scheduling $\sigma$}
While large values of $\sigma$ result in a smoother optimization landscape and better convergence, it also can lead to imprecise alignment as it smooths out fine features. 
Hence, similarly to~\cite{nasa}, we schedule $\sigma^{(t)}$ across the course of training via an additive trigonometric scheduler; see~\eq{scheduler} and \Section{results}.

\subsection{Keypoint energy}
\label{sec:keypoint}
The keypoint energy measures the alignment of keypoints:
\begin{equation}
\loss{key}(\pose; \keypoints_a, \keypoints_b) = 
\sum_{\keypoint_a, \keypoint_b}
\| \keypoint_a - (\R \keypoint_b + \t) \|_2^2
\label{eq:keypoint}
\end{equation}
where $\pose(\x) {=} \R \x + \t$ is the relative pose.
The 3D keypoints are computed by randomly rendering two nearby views from the first scene, and asking the user to select at least \textit{three pairs} of corresponding 2D keypoints in image space.
We then triangulate their 3D position by converting the 2D keypoints into rays, and computing the closest distance between ray pairs~\cite[Fig.3]{selfcalibrating}.
We then ask the user to roughly identify the same keypoints on the second scene with the same process, so the sets $\keypoints_a$ and $\keypoints_b$ are in 1-1 correspondence.

\subsection{Sampling}
\label{sec:sampling}
In \eq{robustmatching}, there is a design degree of freedom in the choice of the sampling set~$\activeset$.
Due to the curse of dimensionality, uniformly sampling $\ball(0, r_a)$ is not only wasteful, but could lead the registration to fall into the wrong local minima, especially if the two scene contain multiple objects with similar geometry (i.e. the registration problem is multi-modal).
To address this, we employ a \textit{Metropolis-Hastings} sampling scheme that \textit{iteratively} updates the active set of samples $\activeset^{(t)}$ during the optimization iterations $(t)$.
As we employ the keypoints to identify the object of interest, we bootstrap Metropolis-Hastings with the keypoint locations, that is $\activeset^{(0)} {\leftarrow} \keypoints_a$. Then, every $N{=}20$ iterations, we follow Algorithm~\ref{alg:sampling} to update the active set $\activeset^{(t)}$.

\begin{algorithm}[b]
    \caption{Sampling Algorithm}
    \label{alg:sampling}
    \SetAlgoLined
    \DontPrintSemicolon
    
    $\activeset^{(t)} \leftarrow \activeset^{(t-1)}$\quad\tcp*{Initialization}
    
    $\candidates \leftarrow \activeset^{(t-1)} + \resolution \cdot \mathcal{U}_3[-1, +1]$ \tcp*{Candidates}

    \For  {each candidate, $x \in \candidates$} {

        \uIf{\: $\S_a^\var(\x) \geq \xi_\S$ \textbf{and} \tcp*{surface value}
        \quad$r(\x; \S_a, \S_b, \pose) \leq \xi_r$ \textbf{and} \tcp*{correspondence}
        \quad$d(\x, \activeset^{(t-1)}) \geq \resolution/10$\tcp*[l]{ min distance to others}}
        {
            
            $\activeset^{(t)} \leftarrow \activeset^{(t)} \cup \{x\}$
            
        }
    }
    return $\activeset^{(t)}$
    
\end{algorithm}
\begin{figure}[t]
\centering
\begin{overpic} 
    [width=.99\columnwidth]
    {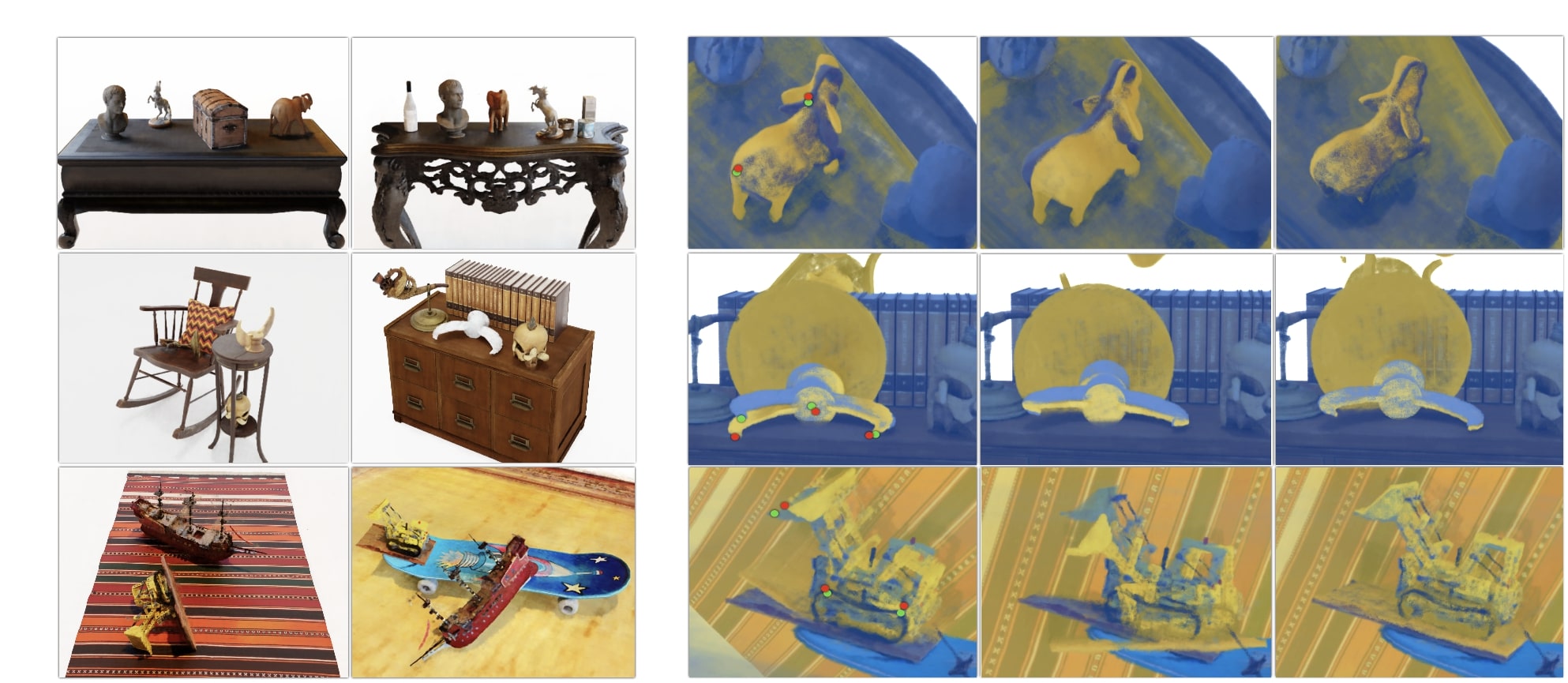}
    \put(6,43){\small Scene A}
    \put(26,43){\small Scene B}
    \put(50,43){\small KP}
    \put(68,43){\small FGR}
    \put(84,43){\footnotesize \algoName}
    \put(40.7,30){\footnotesize \rotatebox{90}{elephant}}
    \put(40.7,16){\footnotesize \rotatebox{90}{pedestal}}
    \put(40.7,4){\footnotesize \rotatebox{90}{lego}}
    \put(-0.5,33.5){\tiny \CIRCLE{1}}
    \put(-0.5,20){\tiny \CIRCLE{2}}
    \put(-0.5,6){\tiny \CIRCLE{3}}
\end{overpic}
\\[0.7em]
\resizebox{\linewidth}{!}{ %
\begin{tabular}{@{}cccccccccc@{}}
\multirow{2}{*}{} Object  & \multicolumn{3}{c}{$10^2 \cdot\Delta \mathbf{t}\downarrow$} & \multicolumn{3}{c}{{$ \Delta \mathbf{R}\downarrow$} } & \multicolumn{3}{c}{$10^2 \cdot \text{\textbf{3D-ADD}}\downarrow$} \\
\cline{2-10}
\\[-0.7em]
\: (Pair No.)  &    KP & FGR & Ours  \:&     KP&FGR & Ours \:&   KP & FGR & Ours  \\
\midrule                      

\multicolumn{1}{c}{bust \CIRCLE{1}}   &     7.78  & 7.94 & \textbf{0.73}  \:&  9.93 &   10.08 & \textbf{1.39}  \:&  7.16 &  7.02 & \textbf{0.77} \\

\multicolumn{1}{c}{elephant \CIRCLE{1}}   &     14.01 & 12.96 & \textbf{0.73}  \:&  17.16 &     15.59 & \textbf{1.00}  \:&  13.97 &  12.87 & \textbf{0.76} \\

\multicolumn{1}{c}{horse \CIRCLE{1}}   &     7.88 & 2.10 & \textbf{0.99}  \:&  14.56 &     7.32 & \textbf{1.77}  \:&  6.67 &  2.09& \textbf{0.88} \\

\multicolumn{1}{c}{pip \CIRCLE{2}}   &     4.01 & 18.93 & \textbf{0.25}  \:&  6.40 &     30.01 & \textbf{1.18}  \:&  4.61 &  20.36 & \textbf{0.46} \\

\multicolumn{1}{c}{jar \CIRCLE{2}}   &     10.40 & 7.17 & \textbf{0.18}  \:&  20.46 &     17.03 & \textbf{2.84}  \:&  11.31 &  8.13 & \textbf{0.97} \\

\multicolumn{1}{c}{pedestal \CIRCLE{2}}   &     5.62 & 8.13 & \textbf{0.69}  \:&  15.83 &     11.69 & \textbf{2.42}  \:&  8.42 &  9.51 & \textbf{1.23} \\

\multicolumn{1}{c}{lego \CIRCLE{3}}   &     14.41 & 12.97 & \textbf{2.09}  \:&  38.08 &     86.29 & \textbf{3.89}  \:&  18.75 &  18.40 & \textbf{2.97} \\

\multicolumn{1}{c}{ship \CIRCLE{3}}   &     13.48 & 4.80 & \textbf{0.75}  \:&  20.52 &     10.74 & \textbf{1.35}  \:&  15.89 &  5.36 & \textbf{0.91} \\

\end{tabular}
}
\caption{
\textbf{Rendered scenes} -- 
When registration only relies on human annotations (KP, green/red dots), the alignment has low precision.
Distilling a point cloud by computing NeRF expected depth, and employing Fast Global Registration~(FGR)\cite{fgr} to align the scenes, results in sub-optimal alignment, as the point clouds are noisy.
Registering with \algoName results in accurate alignment, as qualitatively observable in the z-fighting between the two scenes.
} %
\label{fig:rendered_scenes}
\end{figure}

\begin{figure}
\begin{overpic} 
    [width=.99\columnwidth]
    {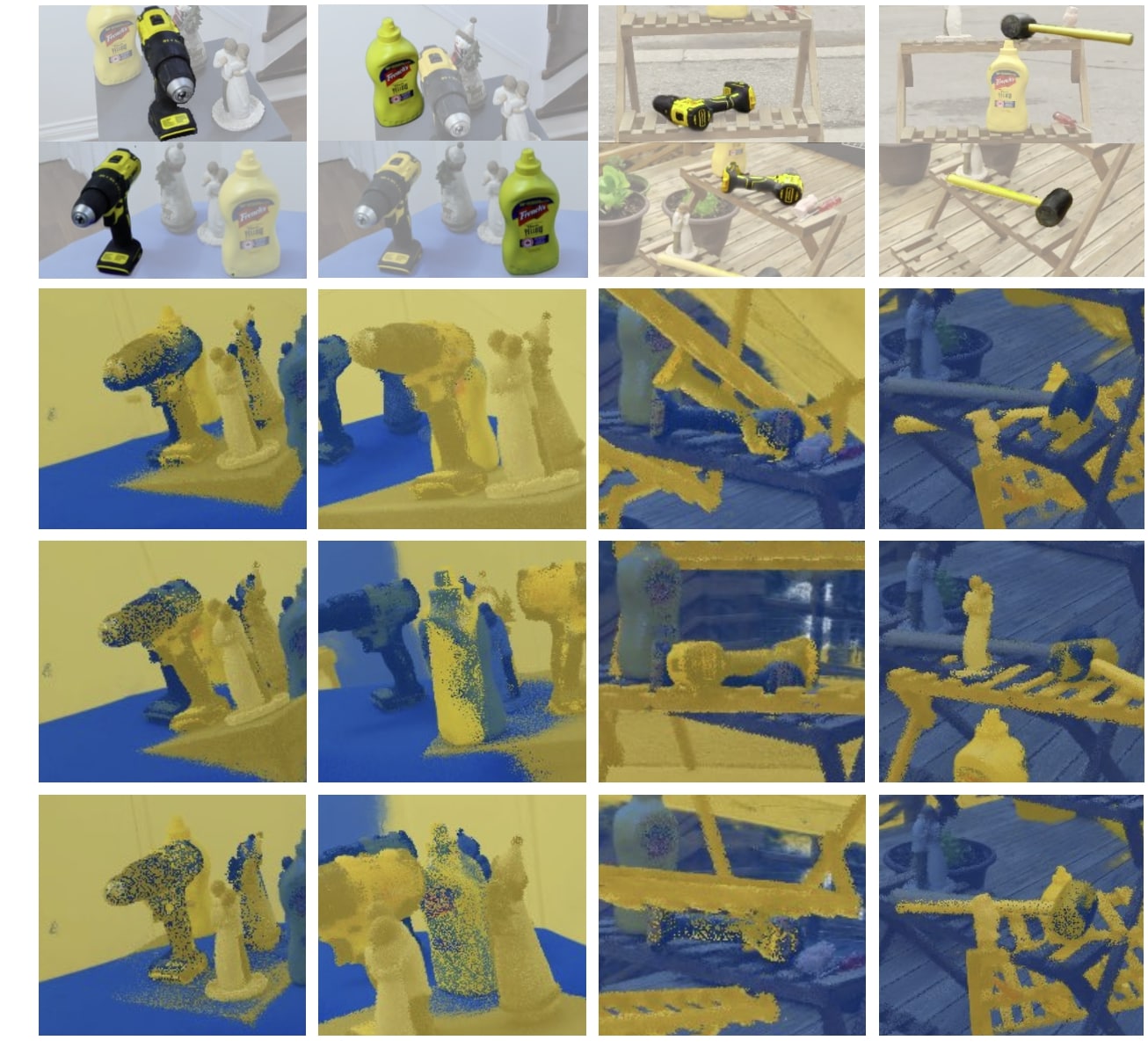}
    \put(0,81.2){\footnotesize \rotatebox{90}{Scene A}}
    \put(0,69){\footnotesize \rotatebox{90}{Scene B}}
    \put(0,56){\footnotesize \rotatebox{90}{KP}}
    \put(0,32.5){\footnotesize \rotatebox{90}{FGR}}
    \put(0,8){\footnotesize \rotatebox{90}{nerf2nerf}}
    \put(13,-0.75){\footnotesize {drill}}
    \put(35,-0.75){\footnotesize {mustard}}
    \put(61,-0.75){\footnotesize {drill}}
    \put(83,-0.75){\footnotesize {hammer}}
\end{overpic}
\caption{
\textbf{Real scenes} -- 
registration of NeRF scenes captured in the wild -- registration target is highlighted for visualization purposes.
Note no ground truth is available for real-scenes, so we are only able to evaluate qualitatively.
We also show the results for keypoint registration~(KP) and Fast Global Registration (FGR)~\cite{fgr}.}
\label{fig:real}
\end{figure}
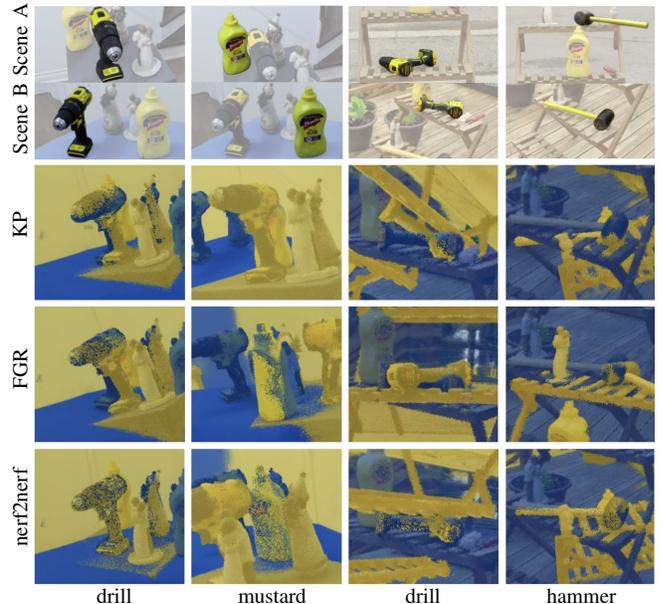

\section{Results}
\label{sec:results}
We evaluate the effectiveness of \algoName compared to point cloud registration when point clouds are extracted from NeRF expected ray terminations~(\Section{exp1}).
Then, we show an application of  \algoName that employs registration to fuse two incomplete NeRF captures into a single one~(\Section{exp2}).
Finally. we conduct a set of ablation studies to justify our design choices~(\Section{exp3}).

\paragraph{Implementation}
The smooth surface field is distilled according to \eq{surfacefielddiscrete} with $\delta{=}0.05$ into an 8 layer, IPE-conditioned MLP of width 256.
The smoothness levels are set to $\var^{(0)}{=}d/5$ and $\var^{(T)}{=}d/10$, where $d{\approx}0.04$ is the mean maximum distance between annotated keypoints on each object.
In our sampler, $\rho {=} r/100$ where r is the radius of the scene and $\xi_S {=} (max_{\x \in \activeset^{(t)}}(\S_a^\var(\x)))/e^2$.
The value of $\xi_r$ is set to the adaptive hyper-parameter $\kernelCutoff$ from the robust kernel.
We use Adam \cite{adam} optimizer for the optimization and learning rate of 0.02 for the rotation parameters, 0.01 for translation parameters and 0.01 for the adaptive kernel parameters, running the optimization for $T{=}10k$ iterations, in under five minutes on an NVIDIA Tesla P100 GPU.
Although this could also be achieved in closed-form by \textit{shape-matching}~\cite{regcourse}, we initialize our solve by optimizing \textit{only} the keypoint energy for $T{=}2k$ iterations.

\subsection{Registration of NeRFs -- \Figure{rendered_scenes} and \Figure{real}}
\label{sec:exp1}
We evaluate the performance of \algoName on multiple objects set in different environments and illumination configurations.
As a baseline for classical methods, we select \textit{Fast Global Registration} (FGR)~\cite{fgr} -- a commonly used algorithm for robust registration of point clouds (we set their voxel size hyper-parameter to 0.001).
The point clouds are extracted from the predicted depth map of NeRF and then the keypoints are used to crop out the object -- we found FGR would completely fail to register otherwise.
Due to the stochastic nature of FGR and \algoName, the best result out of 10 tries are reported.
The metrics for keypoint registration~(KP) are also reported; note this is the starting point of \algoName.

\paragraph{Dataset}
To enable controlled experiments, we introduce a synthetic dataset of three (pairs of) rendered scenes, and train a NeRF model using its classical architecture~\cite{NeRF}.
The scenes are photo-realistically rendered in Blender, with real environment lighting maps and object models from PolyHaven~\cite{polyhaven2021}, with additional object models from NeRF Blender scenes~\cite{NeRF}.
We additionally capture two (pairs of) real-world scenes, which we reconstruct with mipNeRF-360~\cite{mipnerf-360}.
All scenes contain two to three (randomly posed) objects that can be registered between the pair, totalling to $8$ synthetic objects and 6 real objects.

\paragraph{Metrics}
For the rendered scenes, we report the root mean squared error for translation and Euler angles, as well as a modified version of the ADD metric~\cite{ADD}.
Out of the box, ADD cannot be used because we perform registration directly in 3D, and not based on multiple random views of the object.
Specifically, 3D-ADD measures the average distance between corresponding mesh vertices (normalized w.r.t. diameter) when aligned with the predicted pose.

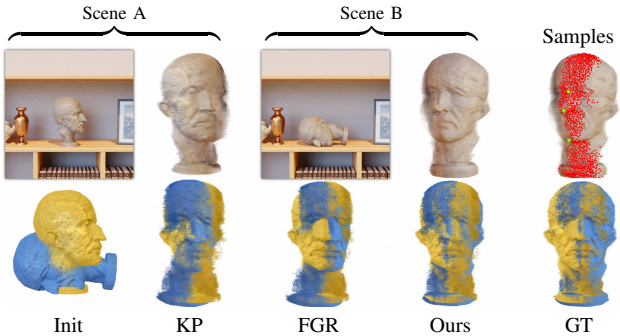
\begin{figure}
\vspace{1.7em}
\begin{overpic} 
[width=.99\columnwidth]
{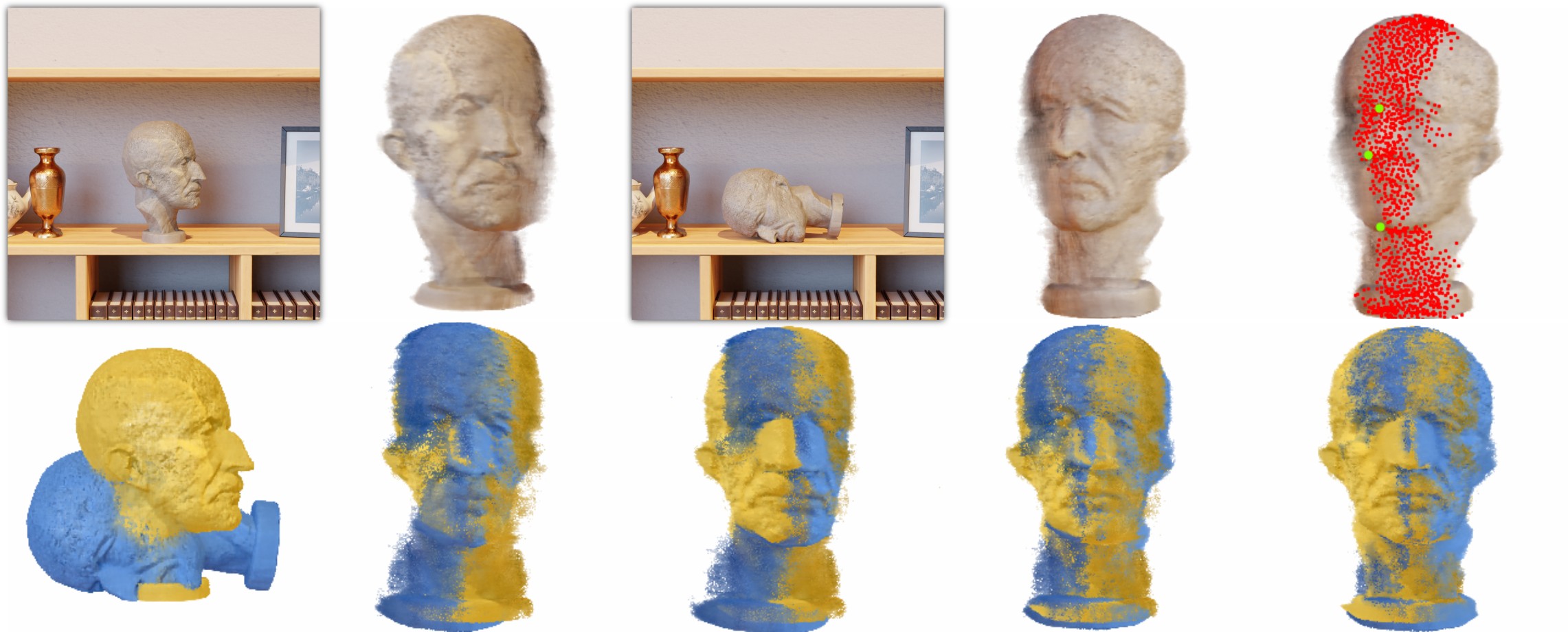}
\put(1,41){$\overbrace{\hspace{8em}}^\text{Scene A}$}
\put(41,41){$\overbrace{\hspace{8em}}^\text{Scene B}$}
\put(84,41){\footnotesize{Samples}}
\put(8,-4){\footnotesize {Init}}
\put(27,-4){\footnotesize {KP}}
\put(46,-4){\footnotesize {FGR}}
\put(66.5,-4){\footnotesize {Ours}}
\put(87.5,-4){\footnotesize {GT}}
\end{overpic}
\vspace{0.8em}
\caption{ 
\textbf{Partial object registration} -- We demonstrate registration of NeRFs trained from \textit{partial} observations of objects where part of the object is unobserved in each scene.
This enables applications such as training a merged NeRF from the combined scenes~\cite{kobayashi2022decomposing}.
}%
\label{fig:nerfusion}
\end{figure}
\subsection{Partial object registration -- \Figure{nerfusion}}
\label{sec:exp2}
We explore the possibility of using our method to fuse two incomplete NeRFs of an object instance (e.g.~due to occlusion) together to make a complete model.
We capture two scenes containing the ``Max Planck'' bust, where due to the object placement, only the left (resp. right) side of the object can be photographed -- only a thin strip in the middle of the face is captured in both scenes.
The NeRFs are trained on the images of the object with the background subtracted and the two resultant NeRFs are registered by \algoName.
Due to the limited overlap between the scenes, we use $\var^{(0)}{=}d/10$ to $\var^{(T)}{=}0$ and smaller learning rates ($.01$ for kernel and $.0005$ for pose) to get a more precise registration.

\begin{figure}[!t]
\centering
\resizebox{\linewidth}{!}{ %

\begin{tabular}{@{}r|cccccc@{}}
& \textbf{\begin{tabular}[c]{@{}c@{}}no $\lambda$ \\ annealing\end{tabular}} &
  \textbf{\begin{tabular}[c]{@{}c@{}}no $\sigma$ \\ annealing\end{tabular}} &
  \textbf{\begin{tabular}[c]{@{}c@{}}uniform \\ sampling\end{tabular}} &
  \textbf{\begin{tabular}[c]{@{}c@{}} density \\ registration \end{tabular}} &
  \textbf{\begin{tabular}[c]{@{}c@{}} radiance \\ registration \end{tabular}} &
  \textbf{Ours} \\ 
\midrule      
\textbf{$10^2 \cdot \Delta$t} $\downarrow$ & $1.56\pm 0.69$ & $3.36\pm 0.90$ & $18.41\pm 8.97$ & $21.13\pm 10.51$ & $6.77\pm 2.36$ & \boldmath$0.95\pm 0.31$ \\
\textbf{$\Delta$R} $\downarrow$ & $2.67\pm 1.27$ & $5.11\pm 1.53$ & $37.81\pm 13.29$ & $39.74\pm 16.81$ & $18.08\pm 5.56$ & \boldmath$1.65\pm 0.59$ \\
$10^2 \cdot$ \textbf{3D-ADD} $\downarrow$ & $1.54\pm 0.68$  & $3.32\pm 0.93$ &  $22.24\pm 7.08$ & $25.18\pm 10.74$ & $8.16\pm 1.84$ & \boldmath$0.95\pm 0.31$
\end{tabular}
} %
\\[.5em]
\begin{overpic} 
[width=.99\columnwidth]
{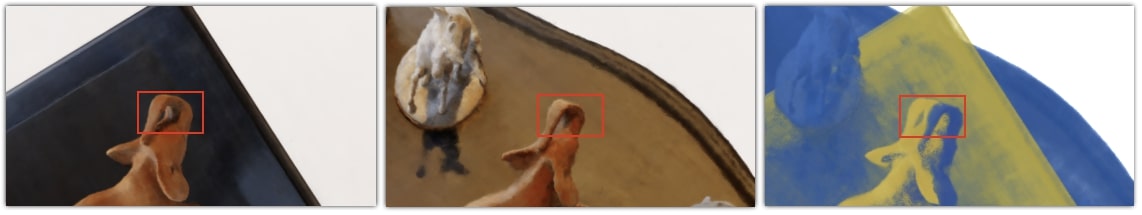}
\put(10,-2.7){\footnotesize{Scene A}}
\put(45,-2.7){\footnotesize{Scene B}}
\put(78,-2.7){\footnotesize{Registration}}
\end{overpic}
\caption{
\textbf{Ablations} -- 
(top) We ablate various features of our method on the three objects in (pair of) Scene~\CIRCLE{1}.
We show the results of the registration in each case averaged over five random restarts.
(bottom) A qualitative example illustrating how differences in scene illumination can cause ``radiance registration'' to fail.
} %
\label{fig:ablations}
\end{figure}

\subsection{Ablations -- \Figure{ablations}}
\label{sec:exp3}
We perform an ablation study on various aspects of our method, reporting the averaged results,
where we find that our proposed formulation achieves the best average registration accuracy among these options.
In more details, we investigate:
whether the annealing of $\lambda$ is necessary, by setting $\lambda {=} 0$, thereby disabling $\loss{key}$;
whether the annealing of $\sigma$ is necessary, by setting $\sigma {=} 0.006$;
whether Metropolis-Hastings sampling is necessary, by instead sampling uniformly over $\ball(0, r_a)$;
whether we can perform registration using density instead, by replacing $\surface$ in the $\loss{match}$ with $\density$;
whether we can perform registration using radiance instead, by replacing $\surface$ in~$\loss{match}$ with $\radiance$.

\section{Conclusions}
Robust pairwise registration is a fundamental tool found in digital processing toolboxes acting on images~\cite{jiang2019linearized, jiang2021cotr}, point clouds~\cite{pomerleau2015registration, fitzgibbon2003robust} and polygonal meshes~\cite{weise2011realtime, dou2016fusion4d, smpl}.
With the emergence of neural fields as a popular representation of 3D scenes~\cite{neural-fields}, \llg{the question arises as to} whether conversion into a classical representation (i.e. a point cloud) is the only way to implement the operation. 
In this paper we demonstrated that this (lossy) conversion is not necessary, and that operating \textit{directly} on neural fields is not only possible, but also performs better than classical pipelines relying on such conversion.
To fulfill this objective, we introduced the concept of \textit{surface fields} as a geometric representation that can be extracted from NeRFs and that is invariant to illumination configurations. 
We then formalized nerf2nerf registration as a robust optimization problem in the ``style'' of ICP~\cite{regcourse}, and thoroughly analyzed its performance on a novel dataset.

\paragraph{Applications}
While we focused on fundamentals, we believe the most exciting directions for future research lie in the applications this tool enables, for example: co-registering a common object embedded in random scenes can enable the modeling of per-scene illumination -- everyday objects can be transformed into consumer-level light probes; ~pairwise registration is at the core of large-scale bundle adjustment pipelines, where registration residuals model pairwise potentials in belief propagation, enabling applications such as scanning a city from thousands of drone captures (unstructured, i.e. without relying on global photo-consistency).

\begin{figure}
\begin{overpic} 
[width=.99\columnwidth]
{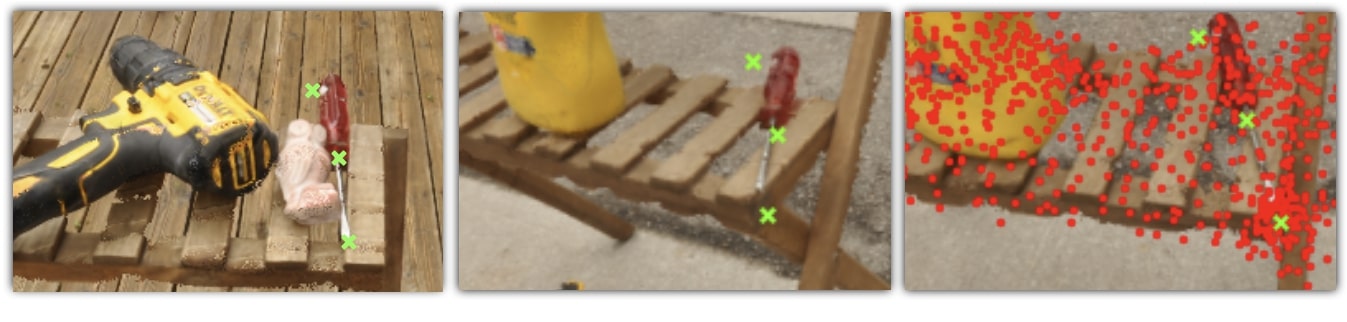}
\put(10,-1.7){\footnotesize{Scene A}}
\put(45,-1.7){\footnotesize{Scene B}}
\put(78,-1.7){\footnotesize{Samples}}
\end{overpic}
\caption{
\textbf{Limitations} -- 
When aligning a small object lying flat on top of a surface our sampler can mistakenly start focusing sampling onto the surface.
This could be resolved by coupling registration of geometry (i.e. surface field) with registration of other properties that are invariant to illumination (e.g. spatial gradients of luminance).
}
\label{fig:limitations}
\end{figure}
\paragraph{Future works}
There are numerous ways to extend our method, from solving the failure case in~\Figure{limitations}, to the implementation of solvers with second order convergence~\cite{pottmann2006convergence}, to techniques that automatically define keypoints rather than relying on user intervention~\cite{gelfand2004slippage}, the integration of ideas from deep-registration~\cite{dcp}, or learning optimal, task specific, field sampling routines~\cite{wang2022deep}.

\clearpage
\renewcommand*{\bibfont}{\small}
\bibliographystyle{IEEEtran}
\bibliography{macros,main}

\end{document}